%% file: main_text.tex
\documentclass[runningheads]{llncs}
\usepackage{graphicx}
\usepackage{soul}
\usepackage{color}

\usepackage{subcaption}
\usepackage{amsmath}
\usepackage{multicol}

\usepackage{hyperref}
\usepackage{graphicx}%
\usepackage{multirow}%
\usepackage{listings}%
\usepackage{caption}
\usepackage{float}
\DeclareMathOperator*{\argmax}{arg\,max}

\begin{document}
\title{Affordance Perception by a Knowledge-Guided Vision-Language Model with Efficient Error Correction\thanks{Supported by TNO ERP APPL.AI program.}}
\titlerunning{Affordance Perception}
\author{G.J. Burghouts, M. Schaaphok, M. van Bekkum, W. Meijer, F. Hillerström, J. van Mil}
\authorrunning{G.J. Burghouts et al.}

\institute{TNO, 2597 AK The Hague, The Netherlands}
\maketitle              % typeset the header of the contribution
\input{sections/abstract}
\input{sections/intro}

\input{sections/relwork}
\input{sections/method}
\input{sections/experiment}
\input{sections/conclusion}
\bibliographystyle{splncs04}
\bibliography{main_text}

\end{document}

%% file: sections/abstract.tex
\begin{abstract}
%The abstract should briefly summarize the contents of the paper in 15--250 words.

Mobile robot platforms will increasingly be tasked with activities that involve grasping and manipulating objects in open world environments. Affordance understanding provides a robot with means to realise its goals and execute its tasks, e.g. to achieve autonomous navigation in unknown buildings where it has to find doors and ways to open these. In order to get actionable suggestions, robots need to be able to distinguish subtle differences between objects, as they may result in different action sequences: doorknobs require grasp and twist, while handlebars require grasp and push.
In this paper, we improve affordance perception for a robot in an open-world setting. Our contribution is threefold: (1) We provide an affordance representation with precise, actionable affordances; (2) We connect this knowledge base to a foundational vision-language models (VLM) and prompt the VLM for a wider variety of new and unseen objects; (3) We apply a human-in-the-loop for corrections on the output of the VLM. The mix of affordance representation, image detection and a human-in-the-loop is effective for a robot to search for objects to achieve its goals. We have demonstrated this in a scenario of finding various doors and the many different ways to open them.

\keywords{Open World Robotics  \and Perception \and Affordance \and Vision-language model \and Knowledge.}
\end{abstract}

%% file: sections/intro.tex
\section{Introduction}\label{sec:intro}

Mobile robot platforms will increasingly be tasked with activities that involve grasping and manipulating objects in open world environments in order to reach a goal \cite{natsuki_2018}. Interacting in such an open world poses challenges for a robot. In order to pursue its goal, the robot needs to take advantage of the actionable possibilities of objects that it encounters, e.g. lift it to get it out of the way, open it to see what is inside, etc. In contrast to a closed world or rigidly structured environment, in an open world the robot needs to be able to adapt to unforeseen events and interact with unknown objects \cite{bessler_2020}. Effective interaction with objects is based on the perception of their affordances \cite{Gaver1991,ardon2021}, what the object offers or provides to an user \cite{gibson_1979}. A button on a door affords pushing, while a handle may also afford pushing \cite{Norman13}. Understanding of affordance allows reasoning about object uses: what possibilities for interaction does the object offer and how can an object be handled and put to use? Understanding affordance comes from the combination of perception and prior knowledge of the world \cite{gibson_1979}. Perception provides clues about possible affordances: what object does the robot see, what properties does it have and what does the object seemingly allow? %Perception will tell us what object we see and what properties they seemingly have.
Combined with prior knowledge this leads to affordance understanding. Potential actions can then be deduced from models that describe how a robot may make use of some property (take action) based on those perceived affordances \cite{ardon2021,bessler_2020}. However, such models do not scale well, because a lot of manual engineering is required to extend them with new object classes and with new contexts for existing object classes. 

In order to generate actionable suggestions, the robot needs to distinguish subtle differences between objects, as they may result in different action sequences: doorknobs require grasp and turn, while door handles require grasp and push. Sources that provide this knowledge about actions may include explicitly structured (semantic) knowledge bases \cite{Speer2012} or e.g. foundational language models like ChatGPT \cite{openai2023gpt4}. Approaches that rely on explicit semantic affordance-action models are precise \cite{ardon2021,bessler_2020}: when one determines an affordance, one can deduce an associated action. This makes for actionable affordances, i.e. affordances with perceivable action possibilities. Similar to explicit knowledge models for perception however, this requires manual engineering and does not scale well. On the other hand, foundational language models scale well and embed a wealth of fine-grained knowledge. They can provide precise information on object affordances, as is demonstrated by ChatGPT when prompted with ``Give me a visual description of a door handle and a door knob and give a step-by-step action list on how to open a door''. The ChatGPT-generated text clearly outlined their visual properties and the steps for using both: ``grasp and push or pull'' for door handles and ``grasp and twist'' for door knobs \cite{chatgpt}. 

Surprisingly, we discover in this paper that foundational vision-language models (VLMs) do not exhibit the same fine-grained distinctions as the language models. VLMs such as GLIP \cite{glip,glipv2} are very flexible and are therefore promising also for affordance perception. They embed knowledge about a wide variety of objects and allow open vocabulary object prompting, which makes their use scalable for new objects and new contexts. However, we find that VLMs models lack the necessary fine-grained semantic differences between objects, such as a doorknob vs. a handlebar, where each requires a different action. This lack of discrimination, makes it difficult to deduce the right action for the robot. To enable the required fine-grained perception, the VLM needs to be fine-tuned. This should require little additional annotations and retraining, because the robot needs to be deployed again quickly to continue its task. %Such knowledge can be induced from properties of the object (discover an affordance from perception), or can be obtained from readily available (prior) knowledge (know an affordance).
We propose a solution by correcting the VLM efficiently for confused objects by a human-in-the-loop.

In this paper, we bring together knowledge representation and foundation models, to achieve a best of both worlds, complemented with a human-in-the-loop that refines missing knowledge. We improve affordance understanding in an open-world setting for a robot whose training and world model includes some, but does not have complete world knowledge. Our contribution is threefold: (1) We provide an affordance representation in a knowledge base that represents precise, actionable affordances for a limited set of relevant objects; (2) We connect this knowledge base to a VLM and prompt it for a wider variety of new and unseen objects similar to those in the knowledge base; (3) We apply a human-in-the-loop to correct the VLM where its finegrained discrimination is lacking.

%% file: sections/relwork.tex
\section{Related Work}\label{sec:related-work}

\subsection{Affordance modelling}
%We want robots to have a meaningful interaction with the world and to be able to interact with the world autonomously. Therefore the robot needs to be able to determine how to reach its goal by manipulating the objects in its environment. 
%The concept of affordances, coined by J.J. Gibson\cite{gibson_1979}, relates the environment and objects relative to its observer. It gives the observer a notion of what action possibilities it has with an object to reach a certain goal. 
There are multiple proposed formalisations of affordances for robotics \cite{sahin_2007} \cite{montesano2007} \cite{cruz2016}, all relating to some extent objects, actions and effects to an agent and its action capability. 
Functional representation of affordances should have the ability for recognition, cognitive and conceptual understanding, how to use and operate an object, invention of new objects/tools for a function \cite{ho2022}. Features/characteristics of objects can be taken into account when detecting affordances, but background knowledge is required \cite{Min_2016}, while affordances may be hidden from current perspective \cite{Min_2016}.
The most common used formalization of affordances is that a potential action exists that could generate the effect, if applied to a specific object in the environment. A single interaction will then produce an instance of this triplet (object, action, effect). The surveys by \cite{aff_survey_2022}, \cite{Zech2017}, \cite{ardon2021}, \cite{Min_2016} provide overviews of the work on affordances for robotics. Affordances can be used for different purposes, such as planning when one knows the desired effects and the objects and one needs to predict the action or for effect prediction when one knows both the object and the action \cite{Graves_2022}.
%They make the distinction between more philsophical approach of affordances and a more practical approach which is required for robotics. They identify target object, action and effects to be the three main components in the different formalisms.
A practical approach for affordances in robotics focuses on affordance templating\cite{Hart2015}, \cite{Hart2022} for more complex manipulation tasks, but this can only handle situation envisioned at design time. Other approaches include deep learning \cite{Chen_2023}, reinforcement learning and value functions \cite{saycan_2022} \cite{Graves_2022}.

\subsection{Affordance Perception}

Most of the affordance perception research, focuses on task-driven object detection \cite{toist,object_use,cotdet}, i.e., finding objects that can cut the tape around a closed box. A context-based Gated Graph Neural Network uses the detected objects in an image and selects the ones that are suitable for a task, given the appearance of an object and the global context of all objects in the scene \cite{object_use}. The category-based approach TOIST extracts the objects from a language description of the task \cite{toist}. The methods in \cite{toist,object_use} leveraged only a limited amount of external knowledge. Therefore, a recent trend is to use large language models (LLMs) to extend the knowledge by a huge amount. In \cite{nocrek} a joint language and vision model was used for object detection. They modeled the detection via language embeddings from Wiktionary object descriptions. The state-of-the-art for affordance perception is the end-to-end learning approach presented in \cite{cotdet}. The approach uses a LLM with chain of thought (CoT) for extracting knowledge about which objects and properties can afford to solve a task. The extracted objects and properties are used to fine-tune an object detection model. This knowledge-conditioned model learns to both detect objects and to recognize visual attributes that provide affordances, e.g., a sharp object can cut. Instead of a comprehensive learning scheme during preparation, our goal is to be flexible in an open world, with new tasks that involve new objects. Therefore, instead of pre-learning for affordance perception, we focus on efficient learning on the job.

\subsection{Open-Vocabulary Object Detection}

For open world robotics, it is key to have open-vocabulary perception models. In computer vision, such methods have rapidly advanced since CLIP \cite{clip}. Here it was shown that contrastive pretraining on a massive number of noisy image-text pairs can result in models with impressive zero-shot generalization capabilities. CLIP is able to classify images, where we are interested in localizing objects in images, a task known as object detection. For object detection, ViLD \cite{vild} and OWL-ViT \cite{vit} extended CLIP with localization capabilities. GLIP \cite{glip} and GLIPv2 \cite{glipv2} followed a different strategy and proposed an architecture and pretraining strategy specifically designed for object detection. Recently GLIPv2 set a new state of the art performance on MS-COCO \cite{coco} for zero-shot tasks, i.e. unseen objects. GLIPv2 performs well at detecting almost any everyday object, therefore we take GLIP as a starting point. To search for the right objects that can provide specific affordances to solve the task at hand (e.g. opening a door via a push bar), we leverage a knowledge base that captures information about affordances and objects. Such knowledge feeds into the model via task-specific textual prompts. Inspired by \cite{ours_icra,ours_iros}, we improve the discrimination by validating known spatial relations between the objects in a scene. One problem with GLIPv2 and similar recent models is their struggle with fine-grained or less common objects or viewpoints \cite{finegrained}. We improve the model's ability to distinguish fine-grained objects by efficient learning on the job.

%% file: sections/method.tex
\section{Method}\label{sec:method}

\subsection{Architecture}\label{sec:architecture}
\begin{figure}[!t]
\includegraphics[width=1.0\textwidth]{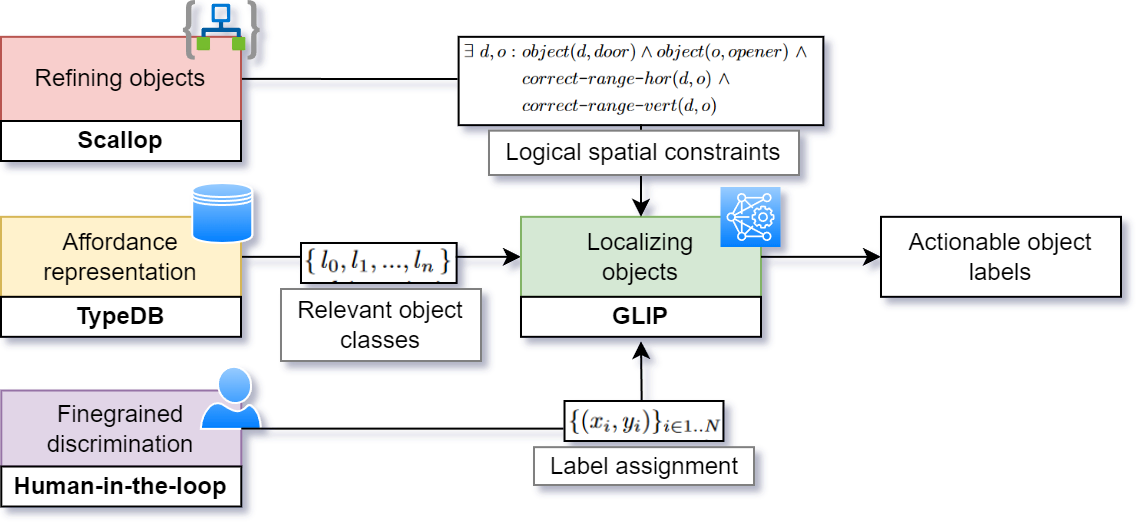}
\centering
\caption{Affordance detection architecture}
\label{fig: affordance-pipeline}
\end{figure}

We propose a modular system as shown in \ref{fig: affordance-pipeline}, consisting of (a) a knowledge base in TypeDB \cite{typedb} for affordance representation, (b) a neuro-symbolic program in Scallop \cite{scallop} for logical constraints, (c) a GLIP module for localizing objects and (d) a human-in-the-loop for label refinement.  

The knowledge base provides labels for relevant object classes to the object detection module, the neuro-symbolic program provides (spatial) constraints on localization of objects and the human-in-the-loop will provide label refinement and correction on identified objects. The output of the object detection is a set of actionable object labels.

% \newpage

\subsection{Affordance representation}
\begin{figure}
    \centering
    \includegraphics[width=0.95\linewidth]{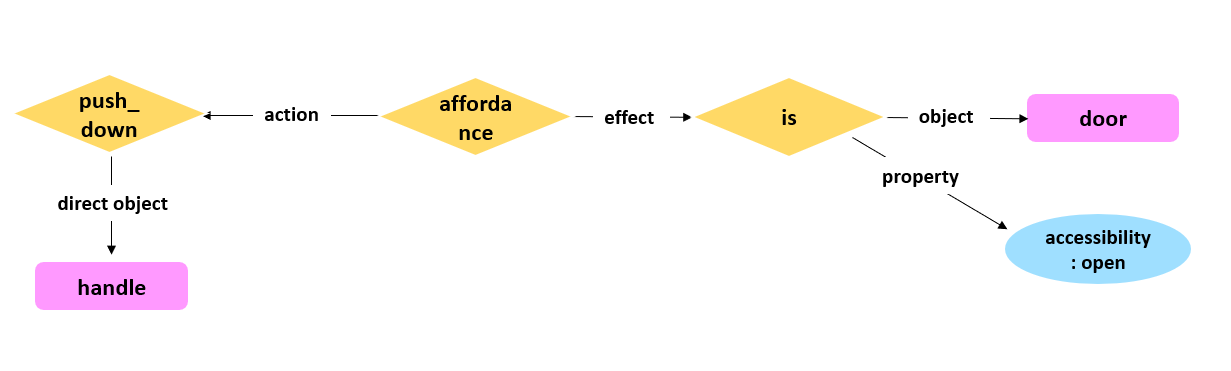}
    \caption{Basic affordance representation using three relations, effect relation, affordance relation and action relation}
    \label{fig:aff_repres_basic}
\end{figure}

To find objects that offer the desired affordance to reach the specified goal, a proper representation of affordances is required. We present a representation of affordances using three relations: the effect relation, the affordance relation and the action relation. The effect relation connects a property to an object. The affordance relation connects an action to an effect and the action relation relates to a direct object. Figure \ref{fig:aff_repres_basic} shows an example of this representation, where the effect is a relation between the object `door' and the property `accessibility', the affordance is the relation between the action `push down' and the effect and the action relation `push down' is related to the direct object `handle'. 

\begin{figure}
    \centering
    \includegraphics[width=0.95\linewidth]{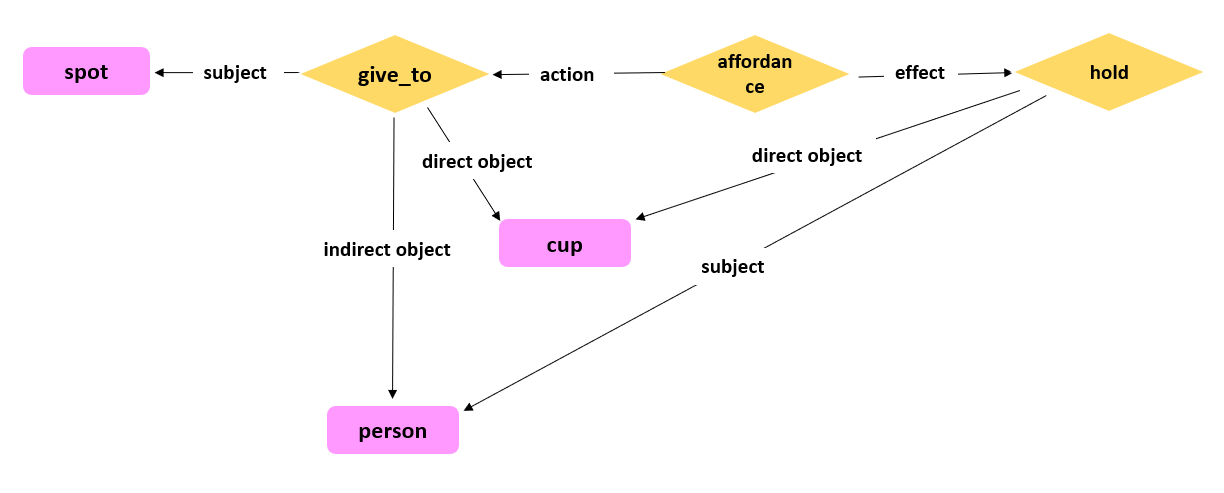}
    \caption{Affordance representation, where the robot SPOT gives a cup to another person, with the effect that this person holds the cup. }
    \label{fig:aff_repres_adv}
\end{figure}

We designed this representation to provide more flexibility in modelling complex affordances and to achieve more flexible, actionable instructions. The representation allows us to (1) explicitly model whether action and effect are on the same or on different objects, (2) represent more detailed and complex affordance structures, including indirect object and other agents (see Figure \ref{fig:aff_repres_adv}). The model structure also allows for (3) different affordances leading to the same effect, which supports quick identification of different affordances, (4) one action to have multiple effects, where it can distinguish between (a) both effects happening jointly or (b) one of the effects taking place. Finally, we  can model (5) effects as a verb (e.g. the effect is that person A ``holds'' a cup), (6) probabilities of an effect taking place, (7) chains of actions to obtain the desired effect (where the ``action'' relation is replaced by an ``action chain'' relation). The affordance representations are modelled in a TypeDB knowledge graph. 
Manually engineering these graphs would be too labour-intensive and inefficient. We can however create baseline versions of these graphs by semi-automated generation (and then manual curation) based on common-sense knowledge available in LLMs (\cite{kommineni2024human}). In order to avoid scalability problems and inference perfomance issues, we envision creating domain- and/or use-case-specific models (e.g. office environment, industrial site, etc.).  

%The following aspects can be taken into account using the representation:  

%\begin{itemize}
%    \item it models explicitly whether action and effect are on the same or on different objects 
%    \item it can represent more detailed and complex affordance structures including indirect object and other agents (see Figure \ref{fig:aff_repres_adv}). 
%    \item it allows modelling different affordances that can lead to the same effect, which supports quick identification of different affordances
%    \item it allows one action to have multiple effects, where it can distinguish between (a) both effects happening jointly or (b) one of the effects taking place.
%    \item effects can be modelled as an verb (e.g. the effect is that person A "holds" a cup) 
%    \item it can include probabilities on how certain it is that an effect will take place
%    \item it allows to model a chain of actions that is required to obtain the desired effect, where the "action" relation is replaced by an "action chain" relation.
%\end{itemize}

%As described in Section \ref{sec:architecture} there are two knowledge bases modelled, one generic affordance knowledge base describing the standard relations between objects, actions and effects, and one situational affordance knowledge base which includes the specific affordances for the objects in the current environment. 

\subsection{Localizing Objects with Affordances}

The goal is to find objects that offer the desired affordance. For instance, if the robot is tasked to open a door, to localize a push bar or handle in an image. There can be multiple means to open doors, e.g. a push bar, a handle, a knob, a button. The set of relevant objects $\{\,l_0, l_1, ..., l_n\,\}$ is taken from a knowledge base, as proposed in \cite{ours_icra}. To localize these objects in image, we prompt the GLIP model \cite{glip} in the string format as proposed in \cite{glip,glipv2}: 

\begin{equation}
    \texttt{<}l_0\texttt{>. }\texttt{<}l_1\texttt{>. }\cdots\texttt{<}l_n\texttt{>}
\end{equation}

\begin{figure}[!h]
    \centering
     \begin{subfigure}[b]{\textwidth}
         \centering
         \includegraphics[width=\textwidth]{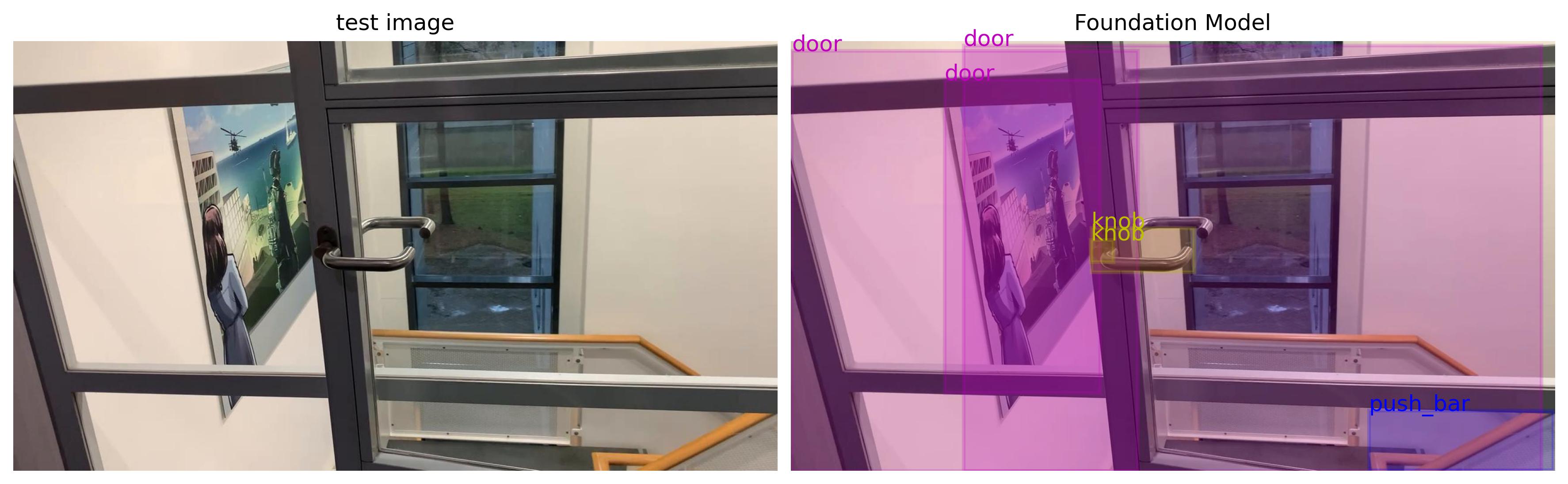}
         % \caption{}
     \end{subfigure}\\
     \centering
     \begin{subfigure}[b]{\textwidth}
         \centering
         \includegraphics[width=\textwidth]{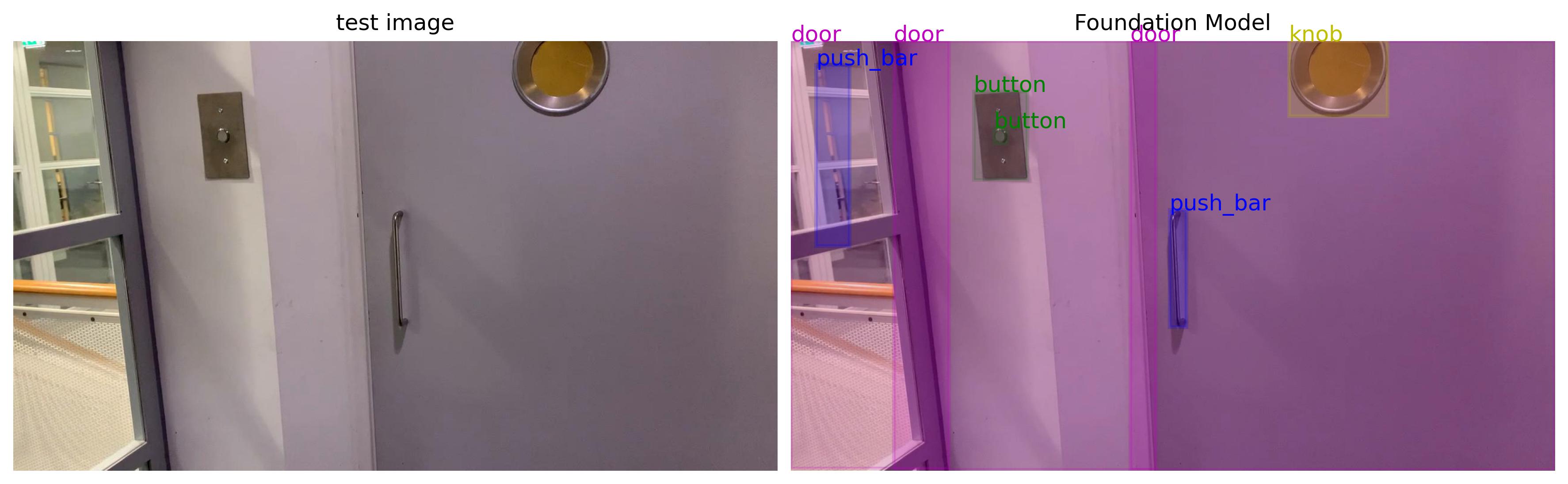}
         % \caption{}
     \end{subfigure}\\
     \centering
     \begin{subfigure}[b]{\textwidth}
         \centering
         \includegraphics[width=\textwidth]{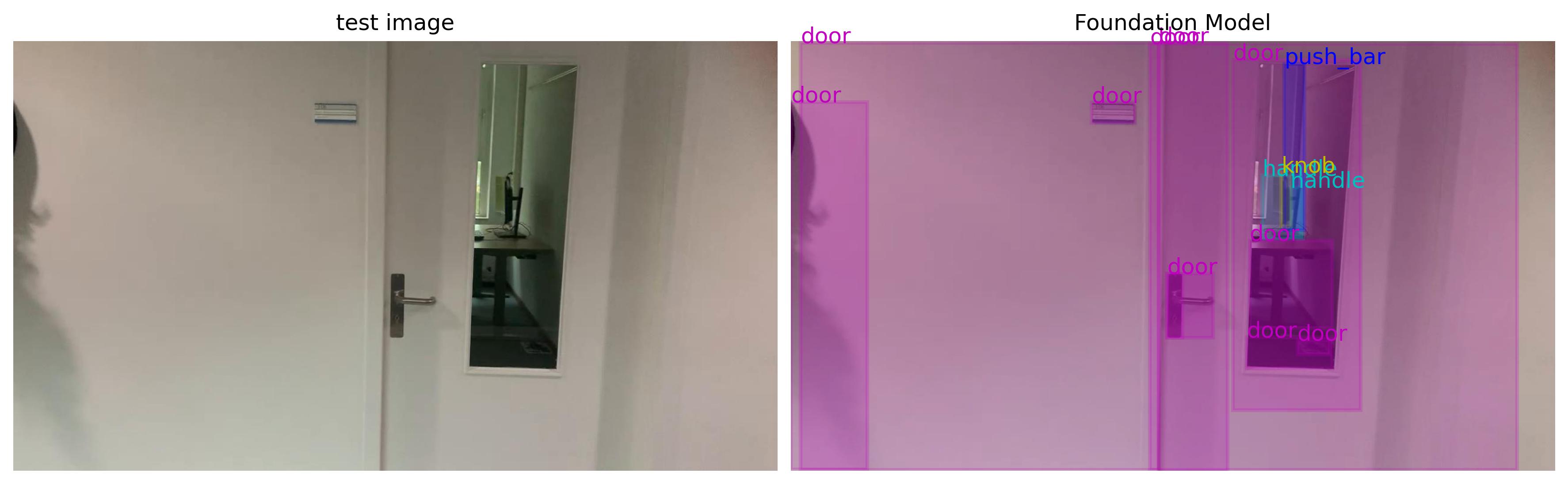}
         % \caption{}
     \end{subfigure}
     \caption{Standard GLIP is incapable of fine-grained discrimination of various door openers.}
     \label{fig:finegrained}
\end{figure}

In accordance with \cite{finegrained}, the model is not able to distinguish between fine-grained object labels. In Figure \ref{fig:finegrained}, results are shown for three different scenes with various door and openers. for example, the door handle is mislabeled as a knob. For a robot, this mistake is crucial, because it will select the wrong action to open it. A knob should be turned, whereas the handle should be pushed downwards. There are many false positives too, respectively the handrail is mistaken for a push bar, the window on the side is mistaken for a knob, and the window frame in the background is mistaken for a door handle. 

\subsection{Finegrained Discrimination by Sparse Human Feedback}\label{sec:relabel}

When using an open-vocabulary detection model, our objective is to increase the model's discrimination between fine-grained objects. Given that most of the relevant objects can be detected (see Figure \ref{fig:finegrained}), the recall is promising already. The problem is mostly in the classification of the objects, i.e. the labels assigned to them (e.g. a handle that is labeled as a knob). To improve the label assignment, our proposal is to involve a human in the process, to reassign labels in an efficient manner. In our scenario, the robot is tasked to navigate through an unknown building. During a first run, it is able to find the relevant objects (recall), but with wrong labels assigned to them. Detected objects are visualized and grouped on a 2D canvas to create an overview, where the human can quickly identify the classes of objects and label a few characteristic instances of each class. In our approach, each object is represented by a feature vector by a visual encoder, we have used CLIP \cite{clip}. The objects can now be distributed on the 2D canvas by a dimensionality reduction, we have used t-SNE \cite{tsne}. This canvas enables the human to quickly assign labels to the encountered object classes.

\begin{figure}[!h]
    \centering
    \includegraphics[width=\textwidth]{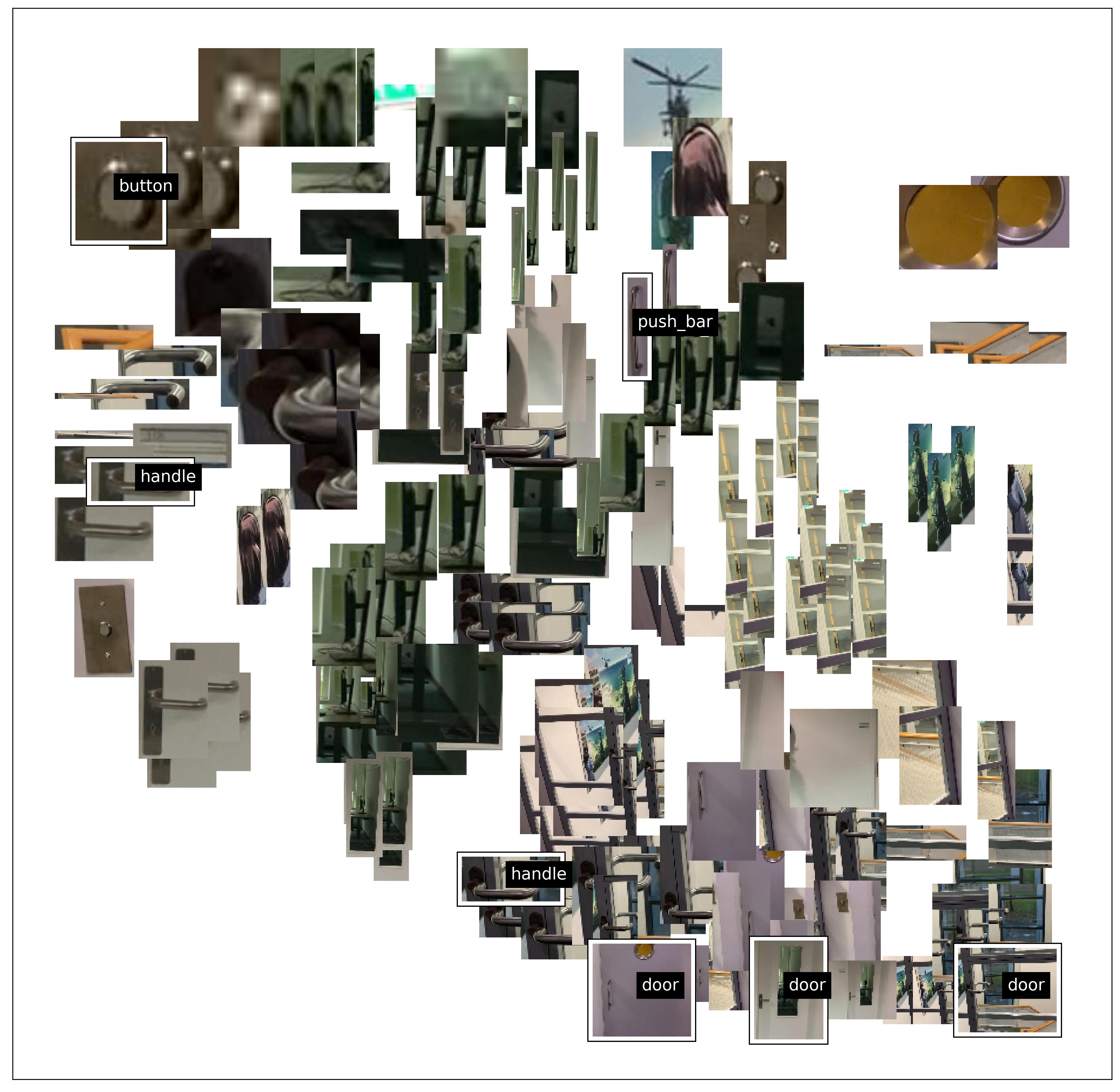}
    \caption{In the overview of all detected objects, the main classes can be identified quickly, as shown by the new labels that were assigned to them by a user.}
    \label{fig:tsne}
\end{figure}

Being provided with a few labels, a task-specific model can now be deployed for the task and environment at hand. For simplicity we leverage a nearest neighbor model, on top of the VLM, to relabel the outputs of the VLM. More specifically, only the labels and confidences are changed by this second model; the boxes remain unchanged. The relabelling model is constructed as follows. The $N$ labels are defined as $D = \{(x_i, y_i)\}_{i \in 1..N}$, where $x_i$ denotes the feature vector of object $i$ and $y_i$ is the provided label for object $i$. An object seen during testing is encoded by feature vector $x_j^{test}$. It gets a label assigned, $y_j^{test}$, with a confidence value $c_j^{test}$. The confidence value $c$ is important for the robot's planning, because it can go to the most confident object first. The assignment of label $y$ and confidence $c$ is inferred the minimal cosine distance to labeled objects:

\begin{equation}
    c(x_j^{test}, x_i) = \frac{x_j^{test} \cdot x_i}{\lVert x_j^{test} \rVert \cdot \lVert x_i \rVert}
\end{equation}

\begin{equation}
    y_j^{test} = \argmax_{y_i} \, c(x_j^{test}, x_i), \,\,\, (x_i, y_i) \in D 
\end{equation}

\subsection{Refining Objects by Spatial Reasoning}\label{sec:spatial}

As can be seen from Figure \ref{fig:finegrained}, there are many false positives when using GLIP. The precision of affordance objects can be enhanced by verifying known spatial relations. For instance, the opener of a door is typically nearby the door. We follow the approach from \cite{ours_icra}. A neuro-symbolic program \cite{scallop} takes the often uncertain objects and their labels and confidences $\{(y_j^{test}, c_j^{test})\}_{j \in 1..M}$ and verifies spatial relations that are predefined by the user before the robot starts the mission. The neuro-symbolic program takes the spatial relations in the form of first-order logic:

\begin{equation}\label{eq:logic}
\begin{split}
\exists \,\, d, o: & \,\,object(d, door) \land object(o, opener) \,\, \land\\
                   & \,\,correct\texttt{-}range\texttt{-}hor(d, o) \,\, \land\\
                   & \,\,correct\texttt{-}range\texttt{-}vert(d, o)
\end{split}
\end{equation}

This specifies that a door opener $o$ is expected to be close to the horizontal side of the door $d$ and vertically close to the middle of the door $d$:

\begin{equation}
correct\texttt{-}range\texttt{-}hor(d, o) = 
max(1 - \frac{hor\texttt{-}dist\texttt{-}from\texttt{-}side(o, d)}{width(d)}, \, 0)
\end{equation}

\begin{equation}
correct\texttt{-}range\texttt{-}vert(d, o) = 
max(1 - \frac{vert\texttt{-}dist\texttt{-}from\texttt{-}middle(o, d)}{height(d)}, \, 0)
\end{equation}

The spatial reasoning approach can be a solution for classes of object affordances if the specification expressed in first-order logic can be a fairly accurate representation of real world conditions. In case there is a higher degree of uncertainty in spatial relations, the approach may be less effective and other formalisms (e.g. possibilistic logic) may yield better results.   

%% file: sections/experiment.tex
\section{Experiment}\label{sec:experiment}

%\hl{Gertjan,  Marianne}: Beschrijving experiment

%\textit{
%Analysis:
%\begin{itemize}
%    \item Using foundational models only is problematic
%    \item Using HitL for actionable labels is improvement
%    \item Using logic from spatial reasoner to remove false positives (Hybrid AI) is further improvement
%\end{itemize}
%}

\subsection{Setup and Dataset}

For experimentation, we consider the scenario where a robot navigates through an office building, and the robot needs to localize the openers such that it can open doors to move through them. In order to test the validity of our approach, we have conducted a limited experiment in a single environment. Videos of three different scenes with various doors and openers were collected for training, comprising 1553 frames. Only 3 frames were labeled by a human user. These frames result from the human feedback as shown in Figure \ref{fig:tsne}. This resulted in 3 labeled doors, 2 labeled door handles, 1 labeled push bar and 1 labeled button. For testing, another video was recorded with similar doors and openers, but under different circumstances such as changed viewpoint and camera distance. This yielded 1370 test frames.

\subsection{Visualization}

\begin{figure}[!h]
    \centering
     \begin{subfigure}[b]{\textwidth}
         \centering
         \includegraphics[width=\textwidth]{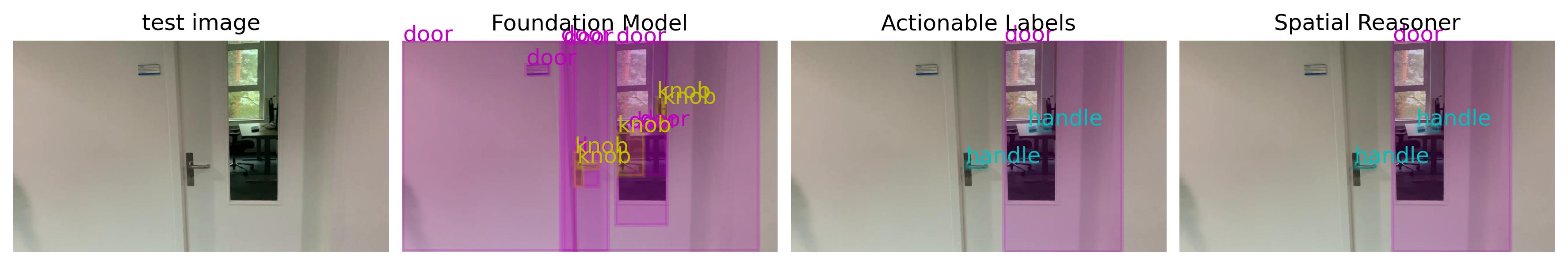}
         % \caption{}
     \end{subfigure}\\
     \centering
     \begin{subfigure}[b]{\textwidth}
         \centering
         \includegraphics[width=\textwidth]{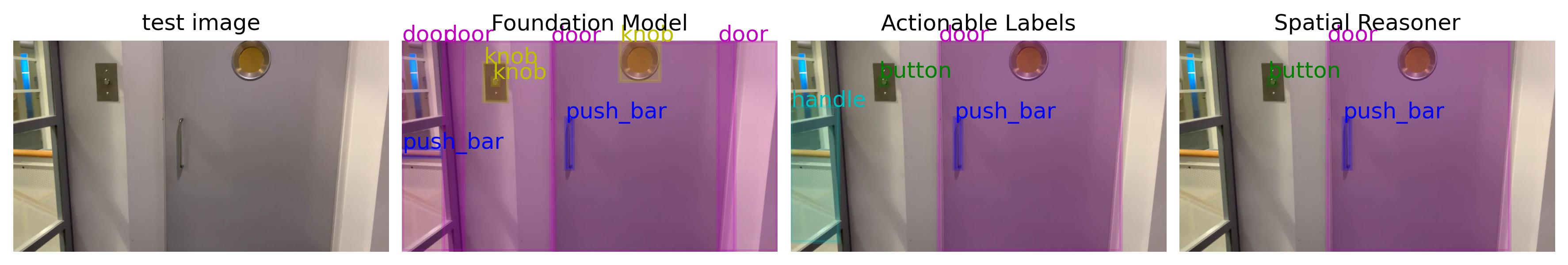}
         % \caption{}
     \end{subfigure}\\
     \centering
     \begin{subfigure}[b]{\textwidth}
         \centering
         \includegraphics[width=\textwidth]{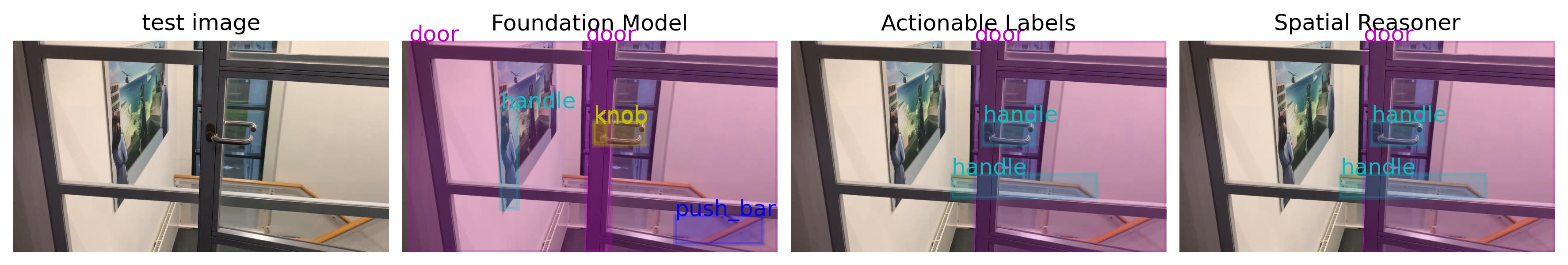}
         % \caption{}
     \end{subfigure}
     \caption{Our relabeling method yields actionable labels.}
     \label{fig:relabel}
\end{figure}

We already established that GLIP's object predictions are not actionable due to many wrong labels and false positives (see Figure \ref{fig:finegrained}). With our relabeling method from Section \ref{sec:relabel}, the predictions improve significantly. Figure \ref{fig:relabel} shows the improved results. In the top row, it shows that the handle was initially labeled as a knob, but it is corrected, and now labeled as a handle. This enables a robot to select the correct action (push down) instead of the wrong action (rotate). In the bottom row, a similar improvement is observed, for relabeling a knob (error) to a handle (correct). In the middle row, the button was corrected from knob to button, and the false positive (push bar) is removed. Most improvement is due to the relabeling method, whereas the spatial reasoning adds minor improvements. For instance, in the middle row, a false positive (handle) is removed, because its spatial relation to the door is not conforming to the expectation (Equation \ref{eq:logic}). A demo clip of our method's predictions on the full test recordings can be found \href{https://youtu.be/jD0fmqcW83Q}{online}.

\subsection{Performance Evaluation}
\begin{figure}[!h]
    \centering
    \includegraphics[width=\textwidth]{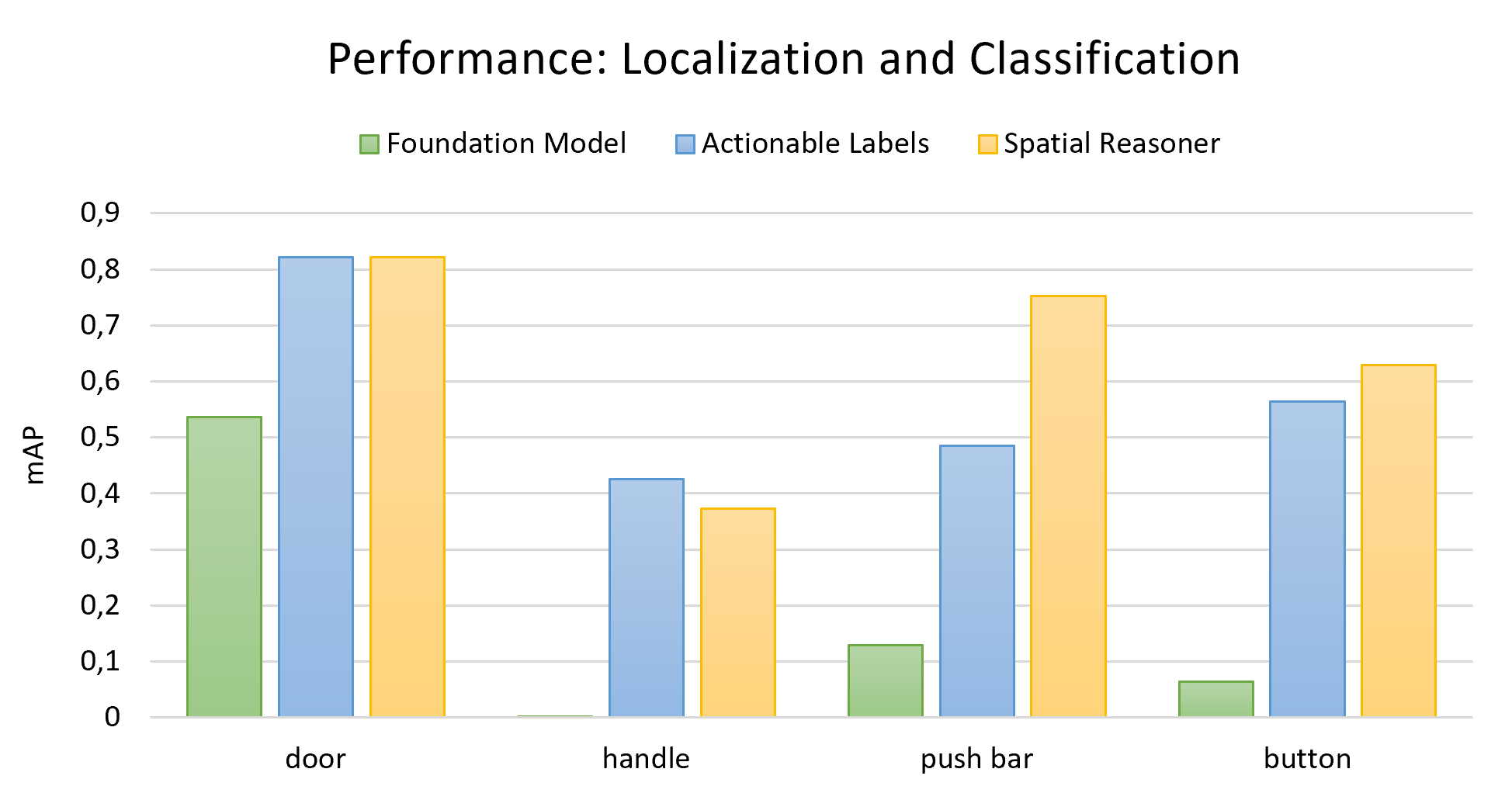}
    \caption{For the localization and classification of affordance objects, the performance of GLIP (green bars) is increased significantly by our method (blue and orange bars).}
    \label{fig:scores}
\end{figure}

A quantitative analysis shows that the performance is indeed improved significantly. The metric for object detection is mAP which measures both the localization and classification accuracy. For the localization, a predicted object should have an overlap with the ground truth of at least IoU $\geq$ 0.5, where IoU is the intersection over union. From the 1370 test frames, 11 images were labeled for validation. The predictions of GLIP are not actionable, as can be observed in Figure \ref{fig:scores} from the low scores for the green bars (left bar in each group), for the door openers, i.e. the handle (almost zero), push bar and button. All scores are below 0.15, which means that less than 1 out of 7 predicted objects is correct. For application in robotics, this is insufficient because the robot would make mostly mistakes while it operates. The performance of our relabeling method is indicated in blue (middle bar in each group). The scores are much higher, showing its effectiveness for finding objects that are related to the affordances of interest. The spatial reasoner is effective for the push bar (orange bar). For this class, GLIP produced many false positives in the background, as there are many elongated shapes that have some resemblance with a push bar. These false positives do not conform to expected spatial relations, i.e. being close to the door. For the other door openers, the spatial reasoning does not impact the predictions much.
 
%% file: sections/conclusion.tex
\section{Conclusion}\label{sec:conclusion}

We have investigated the possibility of affordance analysis by a robot without closed-world assumptions: its training and world model does not include complete world knowledge. Affordances, objects and actions can be precisely modeled in a relational model. A future option could be to automatically extract those from a knowledge base or LLM. Objects from these relations can be visually localized with a VLM. Label assignment quality varies but is overall semantically imprecise (false positives include mix-ups of handrails and push bars) and not actionable (e.g. knob is identified as a handle). Furthermore, detection provides a good starting point for determination of relevant objects in terms of affordances: the recall is good but precision needs improvement. An effective way to determine actionable labels and improve the model, is involvement of sparse feedback from a human who labels characteristic objects from a 2D plot. This drastically improves the mAP (combination of precision and recall for localization and labels) with only a few labels. Our current, limited experiment is a first step towards a more elaborate investigation of our approach in a diverse set of environments and a more open-world setting.  % A mix of explicit modelling, detection and a human-in-the-loop proves to be effective to have a robot search for and analyse objects that allow it to achieve its goals. We demonstrated this in a scenario of finding doors and ways to open them.